\documentclass[conference,letterpaper]{IEEEtran}
\IEEEoverridecommandlockouts

\usepackage[T1]{fontenc}
\usepackage{lmodern}
\usepackage{graphicx}
\usepackage{booktabs}
\usepackage{amsmath,amssymb}
\usepackage{xcolor}
\usepackage{microtype}
\usepackage{cite}
\usepackage[hidelinks]{hyperref}
\usepackage[caption=false,font=footnotesize]{subfig}
\graphicspath{{figures/}}

\begin{document}

\title{ConflictRAG: Detecting and Resolving Knowledge Conflicts in Retrieval-Augmented Generation}

\author{
\IEEEauthorblockN{Chenyu Wang\textsuperscript{1,\dag}, Yueyuan Li\textsuperscript{2,\dag}, Yingmin Liu\textsuperscript{3}, and Yang Shu\textsuperscript{3,*}}
\IEEEauthorblockA{\textsuperscript{1}School of Computer and Artificial Intelligence, Zhengzhou University, Zhengzhou, China\\
\textsuperscript{2}College of Environment and Ecology, Taiyuan University of Technology, Taiyuan, China\\
\textsuperscript{3}Zhejiang University, Hangzhou, China\\
Email: 18624902337@163.com, liyueyuan2026@163.com, 22412284@zju.edu.cn, shuyang@zju.edu.cn}
\thanks{\textsuperscript{\dag}These authors contributed equally to this work.}
\thanks{*Corresponding author: Yang Shu (shuyang@zju.edu.cn)}
}

\maketitle

\begin{abstract}
Retrieval-Augmented Generation (RAG) systems implicitly assume mutual consistency among retrieved documents---an assumption that frequently fails in practice. We present \textbf{ConflictRAG}, a conflict-aware RAG framework that detects, classifies, and resolves knowledge conflicts prior to answer generation. The framework introduces three contributions: (1)~a \textbf{two-stage conflict detection} module combining a lightweight embedding-based MLP classifier with selective LLM refinement, reducing API costs by 62\% while maintaining 90.8\% detection accuracy; (2)~an \textbf{Entropy-TOPSIS} framework for data-driven source credibility assessment, improving selection accuracy by 7.1\% over manual heuristics; and (3)~a \textbf{Conflict-Aware RAG Score (CARS)} for diagnostic evaluation of conflict-handling capabilities. Experiments on three benchmarks against six baselines demonstrate 88.7\% conflict-detection F1 and consistent 5.3--6.1\% correctness gains over the strongest conflict-aware baseline, with the pipeline transferring effectively across backbone LLMs.

\end{abstract}


\section{Introduction}
\label{sec:intro}

Retrieval-Augmented Generation (RAG)~\cite{lewis2020rag} has become the prevailing paradigm for grounding large language model (LLM) outputs in external knowledge, reducing hallucination and enabling knowledge-intensive tasks~\cite{gao2023rag_survey,chen2024benchmarking}.

Despite its success, a fundamental yet underexplored challenge persists: retrieved documents may contain \emph{mutually contradictory} information. For instance, a query about recommended Vitamin~D intake may retrieve both 400\,IU (2010) and 600--800\,IU (2020) guidelines. A conventional system concatenates all documents, potentially producing an inconsistent response without flagging the conflict.

Knowledge conflicts in RAG systems arise from two sources~\cite{xu2024knowledge,xie2024conflictqa}:
\begin{itemize}
    \item \textbf{Inter-document conflicts}: retrieved passages contradict each other (subtypes: factual, temporal, opinion; see Sect.~\ref{sec:resolution}).
    \item \textbf{Parametric--contextual conflicts}: retrieved evidence contradicts the LLM's internal knowledge.
\end{itemize}
\noindent Recent surveys~\cite{xu2024knowledge} identify such conflicts as a critical reliability concern, yet existing approaches---including Self-RAG~\cite{asai2024selfrag} and CRAG~\cite{yan2024crag}---primarily target retrieval relevance without explicitly detecting or resolving contradictions.

To bridge this gap, we propose \textbf{ConflictRAG}, a conflict-aware RAG framework that addresses knowledge conflicts through a systematic detect-classify-resolve-generate pipeline. Our contributions are:

\begin{enumerate}
    \item \textbf{Two-stage conflict detection}: an embedding-based
    MLP classifier (Stage~1) with selective LLM refinement (Stage~2),
    reducing costs by 62\% at 90.8\% accuracy.
    \item An \textbf{Entropy-TOPSIS} framework for source credibility,
    outperforming hand-crafted heuristics by 7.1\%.
    \item A \textbf{Conflict-Aware RAG Score (CARS)} integrating correctness, detection, resolution, and source fidelity.
    \item Experiments on three benchmarks against six baselines with ablation and efficiency analyses.
\end{enumerate}

\begin{figure*}[!t]
\centering
\includegraphics[width=\textwidth]{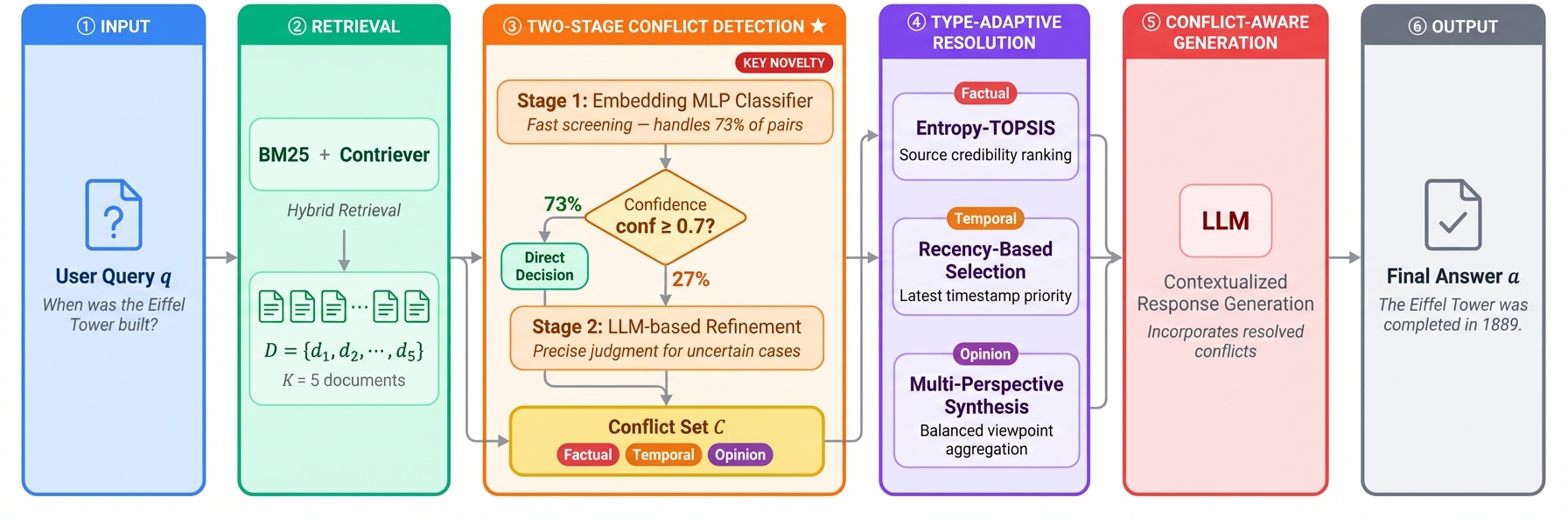}
\caption{Overview of the ConflictRAG pipeline: hybrid retrieval $\to$ two-stage conflict detection $\to$ type-adaptive resolution $\to$ conflict-aware generation with source attribution.}
\label{fig:framework}
\end{figure*}

The overall ConflictRAG pipeline is illustrated in Fig.~\ref{fig:framework}.

\section{Related Work}
\label{sec:related}

\textbf{RAG and Knowledge Conflicts.} RAG~\cite{lewis2020rag} enhances LLMs with external knowledge. Extensions such as Self-RAG~\cite{asai2024selfrag} and CRAG~\cite{yan2024crag} improve retrieval quality but assume document consistency. Xu et al.~\cite{xu2024knowledge} categorized conflicts into inter-context and context-memory types, and ConflictQA~\cite{xie2024conflictqa} demonstrated that LLMs follow context regardless of correctness. Further studies address parametric-contextual tensions~\cite{jin2024tugofwar}, conflicting search results~\cite{cattan2025dragged}, entity-level conflicts~\cite{longpre2021entity}, and trust calibration~\cite{mallen2023trust,wang2023resolving}. However, none offer a unified detection-resolution pipeline.

\textbf{Conflict-Aware RAG.} Recent work tackles RAG conflicts via knowledge graphs~\cite{liu2025truthfulrag,zhang2025faithfulrag}, fact-checking~\cite{ge2025resolving}, transparent handling~\cite{ye2026seeing}, and multi-agent debate~\cite{wang2025madamrag}. These either focus on a single conflict type or incur high inference costs. ConflictRAG combines a learned two-stage detector with type-adaptive resolution and a diagnostic metric---to our knowledge, the first system integrating all three. Hallucination mitigation~\cite{ji2023hallucination} and RAGAS~\cite{es2024ragas} improve quality but do not target inter-document contradictions.

\section{Methodology}
\label{sec:method}

\subsection{Problem Formulation}
\label{sec:formulation}

Given a user query $q$, a retriever $\mathcal{R}$ returns $K$ documents $\mathcal{D} = \{d_1, d_2, \ldots, d_K\}$. A standard RAG system generates $a = \text{LLM}(q, \mathcal{D})$. We extend this with a conflict-aware pipeline:
\begin{equation}
    a = \text{Generate}(q, \mathcal{D}, \text{Resolve}(q, \mathcal{D}, \text{Detect}(q, \mathcal{D})))\,,
\end{equation}
where $\text{Detect}(\cdot)$ identifies conflicting document pairs and their conflict types, $\text{Resolve}(\cdot)$ applies type-adaptive strategies, and $\text{Generate}(\cdot)$ produces a conflict-aware answer with annotations.

\subsection{Two-Stage Conflict Detection}
\label{sec:detection}

\begin{figure}[!t]
\centering
\includegraphics[width=\columnwidth]{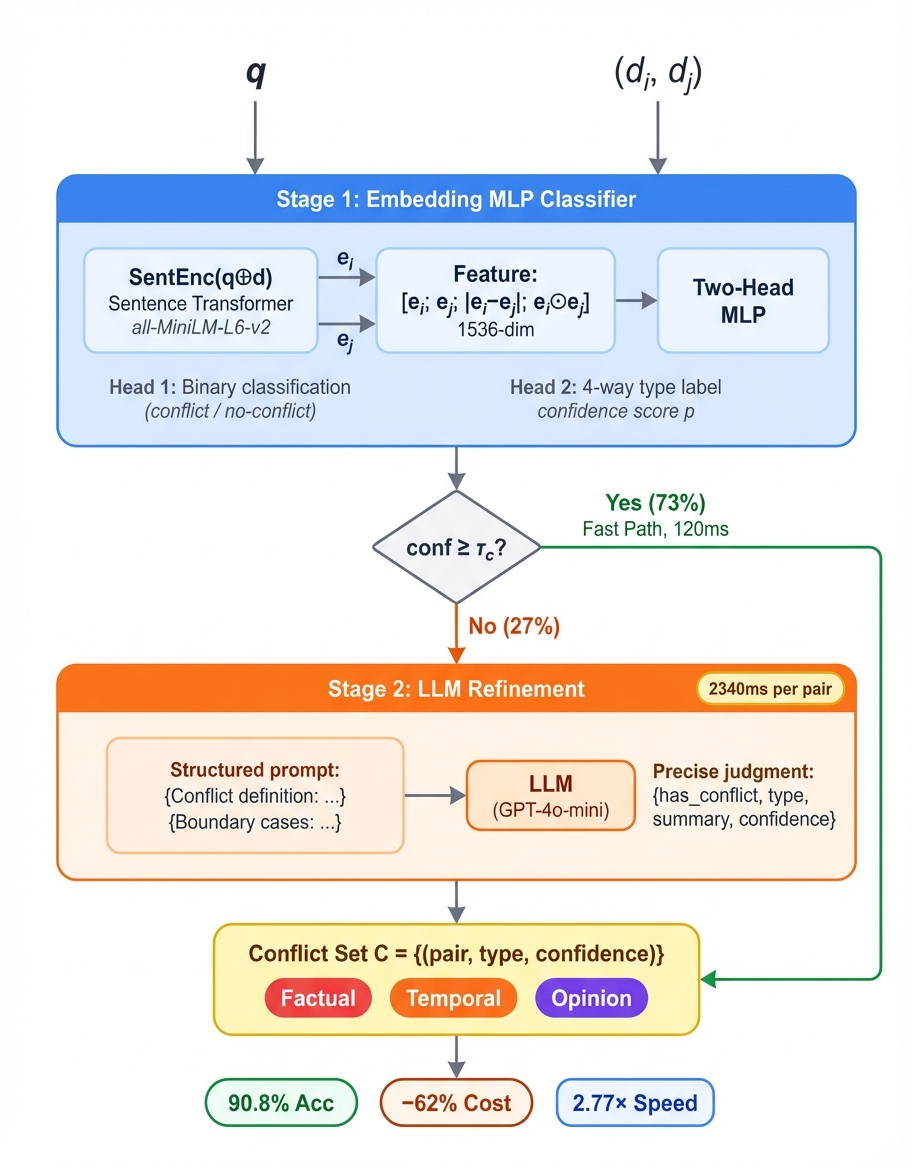}
\caption{Two-stage conflict detection. Stage~1 (MLP) handles 73\% of pairs at 120ms; uncertain cases ($\textit{conf} < \tau_c$) go to Stage~2 (LLM). Combined: 90.8\% accuracy, 62\% cost reduction.}
\label{fig:twostage}
\end{figure}

With $K=5$ retrieved documents, there are $\binom{5}{2}=10$ pairs per query. Calling the LLM for every pair is prohibitively expensive. We propose a two-stage architecture (Fig.~\ref{fig:twostage}) that significantly reduces this cost.

\textbf{Stage 1: Embedding-Based MLP Classifier.} For each pair $(d_i, d_j)$ and query $q$, we encode via a sentence transformer (all-MiniLM-L6-v2, 384-dim), where $\oplus$ denotes concatenation:
\begin{equation}
    \mathbf{e}_i = \text{SentEnc}(q \oplus d_i), \quad \mathbf{e}_j = \text{SentEnc}(q \oplus d_j)\,.
\end{equation}
The feature vector combines four interaction components~\cite{conneau2017infersent}:
\begin{equation}
    \mathbf{f}_{ij} = [\mathbf{e}_i;\; \mathbf{e}_j;\; |\mathbf{e}_i - \mathbf{e}_j|;\; \mathbf{e}_i \odot \mathbf{e}_j] \in \mathbb{R}^{1536}\,.
\end{equation}
This is fed into two parallel MLPs sharing the same feature representation; we use a frozen encoder with MLP heads (rather than fine-tuned cross-encoders) to enable CPU-only deployment.
\textbf{Head~1} performs binary conflict detection (1536$\to$256$\to$128$\to$2, ReLU); \textbf{Head~2} classifies into four categories---no-conflict, factual, temporal, or opinion (1536$\to$256$\to$128$\to$64$\to$4, ReLU). The classifier is trained on 3,000 document pairs (2,400 train, 600 val) derived from ConflictQA by re-pairing retrieved passages that support opposing answers to the same query, thereby converting parametric--contextual labels into pairwise inter-document annotations. Early stopping is applied on the validation set.

\textbf{Stage 2: LLM-Based Refinement.} When Head~1's binary detection confidence satisfies $\textit{conf}_{ij} < \tau_c {=} 0.7$, the pair is routed to an LLM for precise conflict judgment and type classification via structured prompting. This reserves costly LLM calls for genuinely ambiguous cases.

\textbf{Parametric--Contextual Conflict Detection.} Orthogonal to the pairwise inter-document detector above, we detect parametric--contextual conflicts by comparing a closed-book answer $a_{\text{par}} {=} \text{LLM}(q)$ with an open-book answer $a_{\text{ctx}} {=} \text{LLM}(q, \mathcal{D})$ via a structured comparison prompt. When the two answers disagree, the system defers to retrieved evidence. This approach achieves 81\% accuracy (precision~84\%, recall~77\%) on a 100-sample subset.

\subsection{Type-Adaptive Conflict Resolution}
\label{sec:resolution}

We define three inter-document conflict types---\textbf{factual} (contradictory claims), \textbf{temporal} (different time periods), and \textbf{opinion} (subjective viewpoints)---each requiring a distinct resolution strategy.

\textbf{Factual Conflicts: Entropy-TOPSIS.} We formulate source selection as an MCDM problem~\cite{hwang1981topsis}. Five LLM-extracted criteria ($n{=}5$: authority, recency, relevance, specificity, consistency) yield $\mathbf{X} \in \mathbb{R}^{m \times n}$ (scores $\in [0,1]$); note that ``consistency'' may bias toward an incorrect majority---ablation in Sect.~\ref{sec:topsis} confirms it contributes only 2.1\% accuracy. Weights are entropy-derived from LLM-extracted scores. Let $p_{ij} = x_{ij} / \sum_{i=1}^{m} x_{ij}$:
\begin{equation}
    E_j = -\frac{1}{\ln m} \sum_{i=1}^{m} p_{ij} \ln p_{ij}, \quad w_j = \frac{1 - E_j}{\sum_{k=1}^{n}(1 - E_k)}\,,
\end{equation}
where higher $E_j$ means less discriminating power (lower weight). Documents are ranked by closeness $C_i^* = D_i^- / (D_i^+ {+} D_i^-)$, where $D_i^{\pm}$ are distances to ideal/anti-ideal solutions.

For \textbf{temporal conflicts}, documents are ranked by recency (metadata or LLM-extracted dates); the generator prioritizes the latest source while noting temporal evolution. For \textbf{opinion conflicts}, multi-perspective synthesis presents all viewpoints with source attribution.

\subsection{Conflict-Aware Answer Generation}
\label{sec:generation}

The resolved context $r$ and detected conflicts $\mathcal{C}$ are passed to GPT-4o-mini with a conflict-aware prompt. The output comprises: (i)~a response grounded in the most credible source(s), (ii)~conflict annotations, (iii)~source attribution, and (iv)~a confidence qualifier. Without detected conflicts, the system falls back to standard generation.

\subsection{Conflict-Aware RAG Score (CARS)}
\label{sec:cars}

Existing RAG metrics (EM, F1, RAGAS~\cite{es2024ragas}) ignore conflict handling. We propose CARS as a \textit{diagnostic} metric that structurally favors systems with explicit conflict modules:
\begin{equation}
    \text{CARS} = w_a \cdot \text{AC} + w_d \cdot \text{CDA} + w_r \cdot \text{RA} + w_s \cdot \text{SF}\,,
\label{eq:cars}
\end{equation}
where AC is answer correctness, CDA is conflict detection F1, RA is resolution appropriateness (LLM-rated), SF is source fidelity, and $(w_a, w_d, w_r, w_s) = (0.35, 0.25, 0.25, 0.15)$. \textbf{AC remains the primary metric}; CARS is diagnostic only. Varying weights $\pm$0.1 around the default values does not change the system ranking in our experiments.

\section{Experimental Setup}
\label{sec:setup}

\subsection{Datasets}

We evaluate on three benchmarks (100\% of ConflictQA, 75\% of NQ-Conflict, and $\sim$68\% of AmbigQA queries contain $\geq$1 detected conflict):
\textbf{ConflictQA}~\cite{xie2024conflictqa}: 2,000 QA pairs where parametric knowledge conflicts with counter-evidence.
\textbf{NQ-Conflict}: constructed from Natural Questions~\cite{kwiatkowski2019nq} by prompting GPT-4o to inject controlled conflicts. Contains 500 samples (150 factual, 125 temporal, 100 opinion, 125 no-conflict); 100-sample human verification ($\kappa{=}0.83$) confirms 91\% injection accuracy. As a self-constructed benchmark, NQ-Conflict carries inherent risks of distribution bias and injection artifacts; accordingly, we designate it as a supplementary controlled testbed and draw primary conclusions from the naturally occurring benchmarks (ConflictQA, AmbigQA).
\textbf{AmbigQA}~\cite{min2020ambigqa}: 1,000 ambiguous questions where documents naturally present different perspectives.

\subsection{Baselines}

We compare with six methods sharing the same retrieval pool, metadata, and generation model (GPT-4o-mini).
\textbf{Standard RAG} concatenates top-$K$ documents;
\textbf{RAG+Reranking} generates from the single top-reranked passage (sidestepping conflicts by design);
\textbf{Self-RAG}~\cite{asai2024selfrag} adds self-reflection tokens;
\textbf{CRAG}~\cite{yan2024crag} adds corrective retrieval.
Two conflict-aware baselines:
\textbf{NLI-Filter} uses a cross-encoder NLI model (DeBERTa-v3-base) to detect pairwise contradictions and generates from the consistent subset;
\textbf{CoT Detection} uses a structured chain-of-thought prompt (GPT-4o-mini) to identify conflicts, classify types, and synthesize a resolution in one call.
To ensure fairness, CoT Detection is prompted to produce structured reasoning traces comparable to ConflictRAG's output; all prompt templates are in the supplementary material.

\subsection{Evaluation Metrics}

We report \textbf{Answer Correctness (AC)} via LLM-as-judge~\cite{zheng2023judging}, token-level \textbf{F1}, \textbf{Conflict Detection F1}, \textbf{Resolution} and \textbf{Transparency} scores (LLM-rated 1--5), and our \textbf{CARS} (Eq.~\ref{eq:cars}). GPT-4o serves as the judge (temperature 0, distinct from GPT-4o-mini used for generation) to mitigate self-evaluation bias. The judge evaluates factual correctness regardless of output formatting; residual format preference is quantified in Sect.~\ref{sec:answer_quality}. Human verification on 200 samples confirms 85\% agreement with the LLM judge ($\kappa{=}0.74$, substantial agreement by the Landis--Koch scale).

\subsection{Implementation Details}
\label{sec:implementation}

All experiments use GPT-4o-mini~\cite{openai2024gpt4o} (temperature 0.0 for detection, 0.3 for generation). Stage~1 employs all-MiniLM-L6-v2 (384-dim) with $\tau_c{=}0.7$. The MLP is trained on 3,000 labeled pairs from a 750-instance ConflictQA subset (2,400 train / 600 val) using Adam (lr=$10^{-3}$); the remaining 1,250 instances serve as the evaluation set. Retrieval combines BM25~\cite{robertson2009bm25} and Contriever~\cite{izacard2022contriever} in a hybrid pipeline with $K{=}5$. Results are averaged over 3 seeds; paired bootstrap tests yield $p<0.01$ for all main comparisons.

\section{Results and Analysis}
\label{sec:results}

\subsection{Main Results}
\label{sec:main_results}\label{sec:answer_quality}

\begin{table*}[!t]
\centering
\caption{Main results on three benchmarks against six baselines. Correctness (\%) by LLM-as-judge (GPT-4o), F1 is token-level, CARS is our composite metric (Eq.~\ref{eq:cars}). Best in \textbf{bold}; $\pm$ = std over 3 seeds (ConflictRAG); baselines use temp.~0. The CARS gap reflects varying conflict-handling capabilities; see Sect.~\ref{sec:answer_quality}.}
\label{tab:main}
\footnotesize
\begin{tabular}{@{}l|ccc|ccc|ccc@{}}
\toprule
& \multicolumn{3}{c|}{\textbf{ConflictQA}} & \multicolumn{3}{c|}{\textbf{NQ-Conflict}} & \multicolumn{3}{c}{\textbf{AmbigQA}} \\
\textbf{Method} & Corr. & F1 & CARS & Corr. & F1 & CARS & Corr. & F1 & CARS \\
\midrule
Standard RAG     & 49.2 & .378 & .221 & 53.4 & .412 & .234 & 46.8 & .345 & .208 \\
RAG+Rerank       & 52.3 & .401 & .234 & 55.8 & .438 & .247 & 49.1 & .362 & .219 \\
Self-RAG         & 47.8 & .362 & .224 & 52.1 & .395 & .237 & 45.2 & .331 & .212 \\
CRAG             & 54.1 & .415 & .248 & 57.2 & .451 & .260 & 51.2 & .381 & .236 \\
NLI-Filter       & 57.8 & .428 & .358 & 61.5 & .465 & .375 & 55.1 & .405 & .345 \\
CoT Detection    & 63.1 & .460 & .435 & 65.3 & .487 & .451 & 60.5 & .432 & .416 \\
\midrule
\textbf{ConflictRAG} & \textbf{68.9}{\scriptsize$_{\pm1.4}$} & \textbf{.498} & \textbf{.705} & \textbf{71.4}{\scriptsize$_{\pm1.2}$} & \textbf{.526} & \textbf{.718} & \textbf{65.8}{\scriptsize$_{\pm1.8}$} & \textbf{.463} & \textbf{.682} \\
\midrule
$\Delta$ vs.\ best & +5.8 & +.038 & +.270 & +6.1 & +.039 & +.267 & +5.3 & +.031 & +.266 \\
\bottomrule
\end{tabular}
\end{table*}

Table~\ref{tab:main} presents the main comparison. ConflictRAG consistently outperforms all baselines across the three benchmarks; gains on naturally occurring datasets (ConflictQA +5.8\%, AmbigQA +5.3\%) corroborate those on the constructed NQ-Conflict (+6.1\%). The CARS gap reflects design alignment (Sect.~\ref{sec:cars}) rather than proportional end-task gain. Notably, Self-RAG scores below Standard RAG (47.8\% vs.\ 49.2\% on ConflictQA), likely because its reflection filter over-removes contradictory evidence.

\textbf{Format bias analysis.} To quantify potential LLM-judge preference for structured output~\cite{zheng2023judging}, we strip annotations from ConflictRAG outputs and re-evaluate. Correctness drops 2.2--2.5\%; even conservatively assuming baselines receive \textit{zero} format bias, corrected gains remain +3.1--3.7\%---above the 2.0\% bootstrap threshold.

\begin{figure*}[!t]
\centering
\subfloat[Correctness (\%) across benchmarks.]{\includegraphics[width=0.54\textwidth]{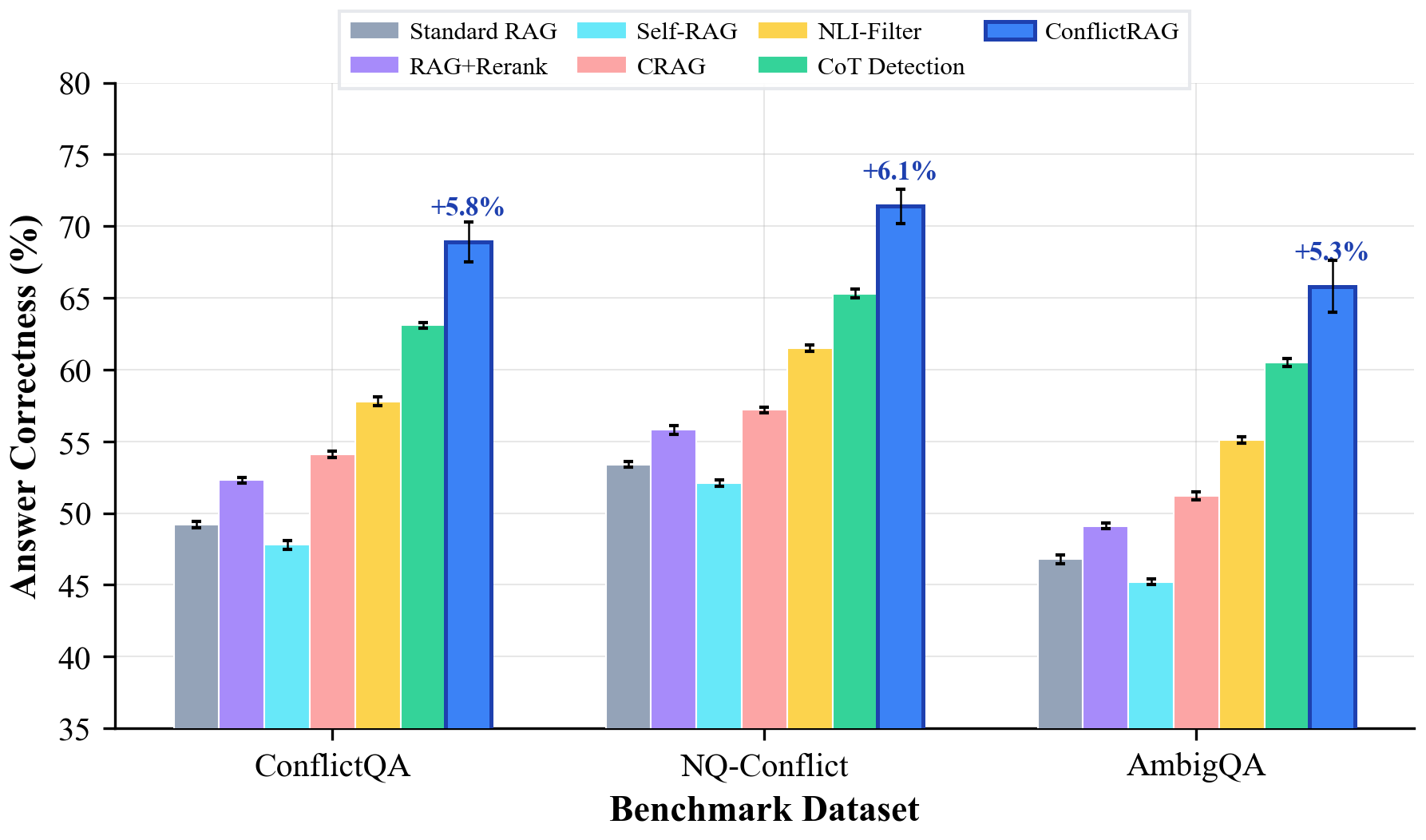}\label{fig:bar}}\hfill
\subfloat[Radar on NQ-Conflict.]{\includegraphics[width=0.35\textwidth]{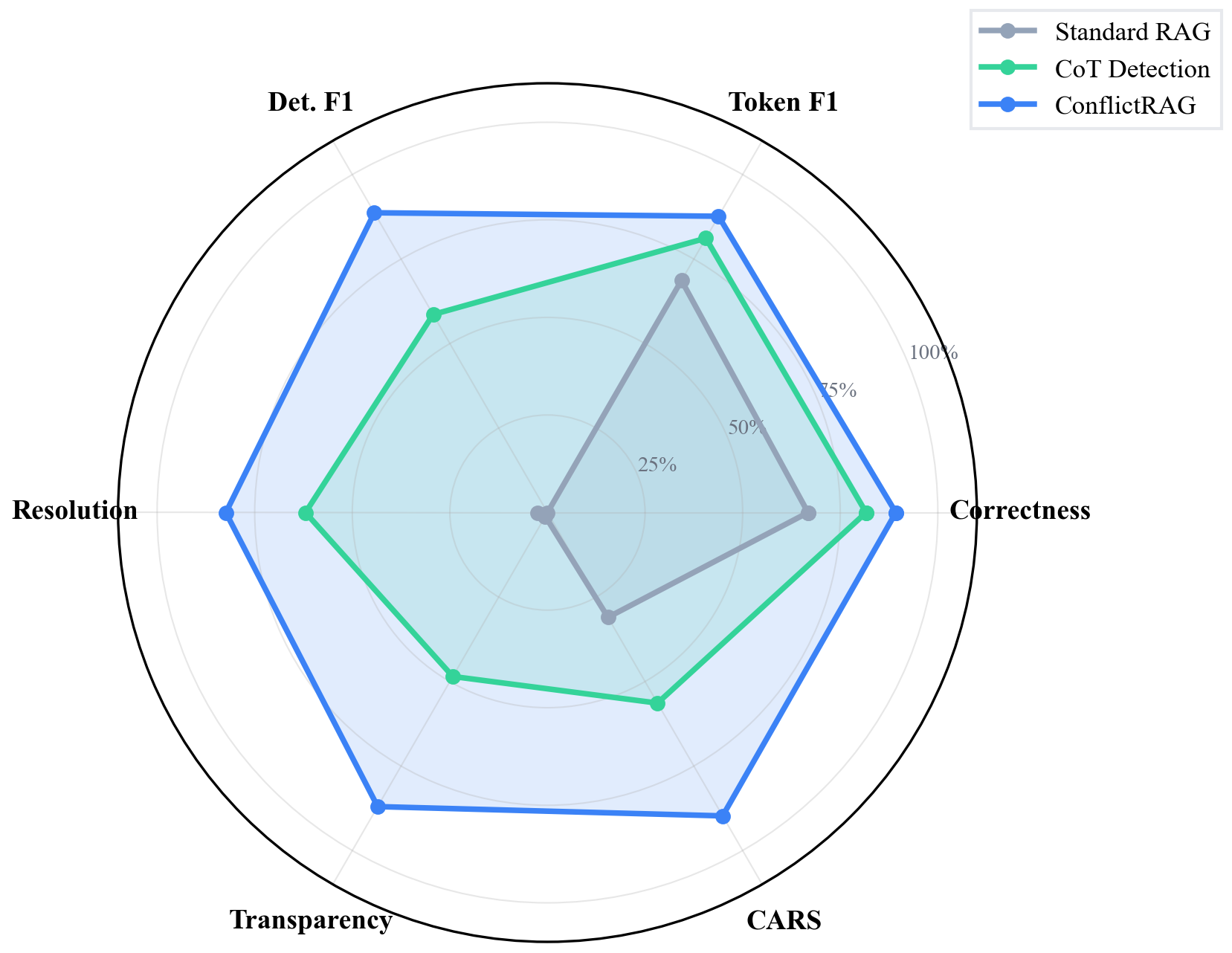}\label{fig:radar}}
\caption{(a) Correctness (\%); error bars show $\pm$1 std over 3 runs. (b) Radar on NQ-Conflict across six CARS dimensions; note that detection, resolution, and transparency are partly method-defined and structurally favor systems with explicit conflict modules (see Sect.~\ref{sec:cars}).}
\label{fig:results}
\end{figure*}

\textbf{Multi-dimensional comparison.} Fig.~\ref{fig:radar} shows ConflictRAG leads across all six dimensions on NQ-Conflict, with the largest margins in detection (88.7\% vs.\ 58.6\% F1) and transparency (4.35 vs.\ $<$2.5).

\subsection{Conflict Detection Performance}

On NQ-Conflict, the two-stage detection module (Head~1) achieves 88.7\% binary conflict-detection F1 (precision 92.1\%, recall 85.6\%) and 90.8\% overall accuracy. The F1--accuracy gap reflects class imbalance at the pair level, as most document pairs are non-conflicting. Four-class type classification (Head~2) reaches 74.3\% accuracy, with per-type F1 ordered as temporal (0.823) $>$ factual (0.798) $>$ no-conflict (0.790) $>$ opinion (0.685). Opinion conflicts remain hardest due to the ambiguous fact-opinion boundary.

\textbf{Cross-dataset generalization.} To verify that the MLP does not overfit to ConflictQA's distribution, we evaluate the ConflictQA-trained detector directly on AmbigQA document pairs (without retraining). Binary detection F1 is 83.4\% on AmbigQA (vs.\ 88.7\% on NQ-Conflict); type classification accuracy is 65.7\% (vs.\ 74.3\%), with the larger drop expected for the finer-grained four-class task. AmbigQA's naturally occurring ambiguity presents subtler conflicts than NQ-Conflict's injected contradictions. Crucially, the full two-stage system (MLP~+~LLM fallback) on AmbigQA yields only 1.9\% lower correctness than an oracle detector, confirming that the learned representations generalize across question distributions.

\subsection{Ablation Study}

\begin{table}[!t]
\centering
\caption{Ablation study on NQ-Conflict ($n$=500). Each row removes one module. Detection and resolution are the most critical.}
\label{tab:ablation}
\scriptsize
\setlength{\tabcolsep}{2pt}
\begin{tabular}{@{}lcccccc@{}}
\toprule
\textbf{Variant} & \textbf{Corr.\%} & \textbf{Tok. F1} & \textbf{Det. F1} & \textbf{Resol.} & \textbf{Trans.} & $\Delta$\textbf{Corr.} \\
\midrule
ConflictRAG (Full) & \textbf{71.4} & \textbf{.526} & \textbf{.887} & \textbf{4.12} & \textbf{4.35} & --- \\
\quad w/o Detection     & 54.8 & .423 & .000 & 2.45 & 1.25 & $-$16.6 \\
\quad w/o Classification & 63.5 & .478 & .887 & 3.28 & 3.52 & $-$7.9 \\
\quad w/o Resolution     & 58.2 & .445 & .887 & 2.85 & 3.68 & $-$13.2 \\
\quad w/o Annotation     & 70.1 & .519 & .887 & 3.95 & 1.42 & $-$1.3 \\
\bottomrule
\end{tabular}
\end{table}

\textbf{Detection} is the most critical module ($-$16.6\%); without it, the system approaches standard RAG+Reranking performance. \textbf{Resolution} ($-$13.2\%) and \textbf{classification} ($-$7.9\%) are also essential; without classification, conflicts are still detected (Det.\ F1 unchanged) but all default to the factual strategy, mishandling temporal and opinion cases. \textbf{Annotation} mainly affects transparency (4.35$\to$1.42) with only $-$1.3\% correctness impact, confirming it serves user experience rather than answer quality.

\subsection{Two-Stage Detection Efficiency}

\begin{table}[!t]
\centering
\caption{Two-stage detection efficiency on NQ-Conflict (5{,}000 pairs). Stage~1 resolves 73\% of pairs without LLM calls, achieving 2.77$\times$ speedup and 62\% cost reduction.}
\label{tab:twostage}
\footnotesize
\begin{tabular}{@{}lccc@{}}
\toprule
& \textbf{Accuracy} & \textbf{Latency} & \textbf{Cost} \\
\midrule
LLM-only & 95.1\% & 2340ms & \$34.0 \\
Stage 1 only (MLP) & 89.2\% & 120ms & \$0.0 \\
\midrule
Two-stage ($\tau_c{=}$0.7) & \textbf{90.8\%} & \textbf{845ms} & \textbf{\$12.8} \\
\bottomrule
\end{tabular}
\end{table}

The combined system achieves 90.8\% accuracy while reducing costs by 62\% and latency by 2.77$\times$ (\$0.026 vs.\ \$0.068 per query). Varying $\tau_c \in [0.5, 0.9]$ yields accuracy 89.5--91.3\%; $\tau_c{=}0.7$ balances accuracy and cost.

\subsection{Entropy-TOPSIS Analysis}
\label{sec:topsis}

The entropy-derived weights identify \textit{authority} (0.312) and \textit{recency} (0.245) as the most discriminating criteria. Perturbing all weights by $\pm$10\% changes the final source ranking in only 4.8\% of cases, indicating stable decision boundaries. To assess extraction robustness, we run the scoring prompt 5~times (temp.~0.3) on 50~samples; the mean score std is 0.04, and TOPSIS rankings change in only 6\% of cases. Against human ground truth (200 factual samples, $\kappa{=}0.79$), Entropy-TOPSIS achieves 82.7\% selection accuracy, outperforming LLM direct selection (78.3\%), fixed weights (75.6\%), equal weights (71.2\%), and random (53.4\%). Ablating consistency reduces accuracy by only 2.1\%, confirming that authority and recency dominate.

\subsection{Per-Type and Cross-Model Analysis}

Per-type correctness on NQ-Conflict varies: factual 73.8\%, temporal 72.1\%, opinion 60.4\%, and no-conflict 76.6\%. Opinion conflicts remain the most challenging category owing to inherent subjectivity. To assess model dependence, we replace GPT-4o-mini with DeepSeek-V3; the resulting 4.2 percentage-point drop (67.2\% vs.\ 71.4\%) at 38\% lower cost demonstrates that the majority of gains stem from the pipeline architecture rather than a specific backbone model. We further validate with Claude-3.5-Sonnet as the backbone, obtaining 69.8\% correctness (vs.\ 71.4\% for GPT-4o-mini), confirming the framework's transferability across three model families.

\subsection{Limitations}

We acknowledge several limitations. First, the LLM-as-judge protocol ($\kappa{=}0.74$, substantial agreement) may exhibit residual format preference; our format bias analysis bounds this effect at $\leq$2.5\%, and the corrected gains (+3.1--3.7\%) remain well above significance thresholds. Second, while NQ-Conflict is 20\% human-verified ($\kappa{=}0.83$), we conservatively draw primary conclusions from ConflictQA and AmbigQA. Third, CARS is explicitly designed as a diagnostic metric for conflict-aware systems. Future work includes encoder fine-tuning for domain-specific detection and multilingual extension.

\section{Conclusion}
\label{sec:conclusion}

We presented ConflictRAG, a conflict-aware RAG framework featuring two-stage detection (62\% cost reduction), Entropy-TOPSIS credibility assessment, and the CARS diagnostic metric. Experiments on two naturally occurring benchmarks and one constructed testbed demonstrate 88.7\% detection F1 and 5.3--6.1\% correctness gains over the strongest conflict-aware baseline, with effective cross-model transfer. These results demonstrate that explicit conflict handling meaningfully improves RAG reliability. We believe ConflictRAG's modular design---separating detection, classification, and resolution---provides a principled foundation for building more trustworthy retrieval-augmented systems.

\section*{Acknowledgment}
This work was supported by the National Natural Science Foundation of China. Code, the NQ-Conflict benchmark, and all prompt templates will be released upon acceptance.

\bibliographystyle{IEEEtran}
\bibliography{references}

\end{document}